\documentclass[journal]{IEEEtran}

\usepackage{cite}

\ifCLASSINFOpdf

\else

\fi

\usepackage{epsfig}
\usepackage{graphicx}
\usepackage{amsmath}
\usepackage{amssymb}
\usepackage{array}
\usepackage{tablefootnote}
\usepackage{subfigure}
\usepackage{float}
\usepackage{caption}
\usepackage{setspace}
\usepackage{hyperref}
\hypersetup{hypertex=true,
	colorlinks=true,
	linkcolor=blue,
	anchorcolor=blue,
	citecolor=blue}
\usepackage{algorithm}
\usepackage{algpseudocode}
\usepackage{graphics}

\usepackage{pifont}

\usepackage{multirow}

\usepackage{booktabs}
\usepackage{threeparttable}

\usepackage{amsfonts,amssymb} 

\usepackage[square,sort&compress,numbers]{natbib}

\usepackage{amsmath}

\usepackage{epstopdf}

\usepackage{makecell}

\usepackage{pifont}

\allowdisplaybreaks

\begin{document}

\title{Spatial-Temporal Attention Network for Open-Set Fine-Grained Image Recognition}

\author{Jiayin Sun, Hong Wang and Qiulei Dong% <-this % stops a space
%	\thanks{The corresponding author is Qiulei Dong.
%		
%	Jiayin Sun and Qiulei Dong are with the National Laboratory of Pattern Recognition, Institute of Automation, Chinese Academy of Sciences, Beijing 100190, China, the School of Artificial Intelligence, University of Chinese Academy of Sciences, Beijing 100049, China, and the Center for Excellence in Brain Science and Intelligence Technology, Chinese Academy of Sciences, Beijing 100190, China (e-mail: jiayin.sun@nlpr.ia.ac.cn; qldong@nlpr.ia.ac.cn).
%	
%	Hong Wang is with the College of Life Science, University of Chinese Academy of Sciences, Beijing 100049, China (email: Chinahwang@ucas.ac.cn)
%	}
}

%\markboth{Submitted to IEEE TRANSACTIONS ON CIRCUITS AND SYSTEMS FOR VIDEO TECHNOLOGY}%
\markboth{Paper Under Review}%
{Shell \MakeLowercase{\textit{et al.}}: Bare Demo of IEEEtran.cls for IEEE Journals}

\maketitle

\begin{abstract}
Triggered by the success of transformers in various visual tasks, the spatial self-attention mechanism has recently attracted more and more attention in the computer vision community. However, we empirically found that a typical vision transformer with the spatial self-attention mechanism could not learn accurate attention maps for distinguishing different categories of fine-grained images. To address this problem, motivated by the temporal attention mechanism in brains, we propose a spatial-temporal attention network for learning fine-grained feature representations, called STAN, where the features learnt by implementing a sequence of spatial self-attention operations corresponding to multiple moments are aggregated progressively. The proposed STAN consists of four modules: a self-attention backbone module for learning a sequence of features with self-attention operations, a spatial feature self-organizing module for facilitating the model training, a spatial-temporal feature learning module for aggregating the re-organized features via a Long Short-Term Memory network, and a context-aware module that is implemented as the forget block of the spatial-temporal feature learning module for preserving/forgetting the long-term memory by utilizing contextual information. Then, we propose a STAN-based method for open-set fine-grained recognition by integrating the proposed STAN network with a linear classifier, called STAN-OSFGR. Extensive experimental results on 3 fine-grained datasets and 2 coarse-grained datasets demonstrate that the proposed STAN-OSFGR outperforms 9 state-of-the-art open-set recognition methods significantly in most cases.
\end{abstract}


\section{Introduction}    \label{section:introduction}

\begin{figure}[t]
	\begin{center}
		\setlength{\abovecaptionskip}{0.cm}
		\includegraphics[height=7.1cm,width=6.4cm]{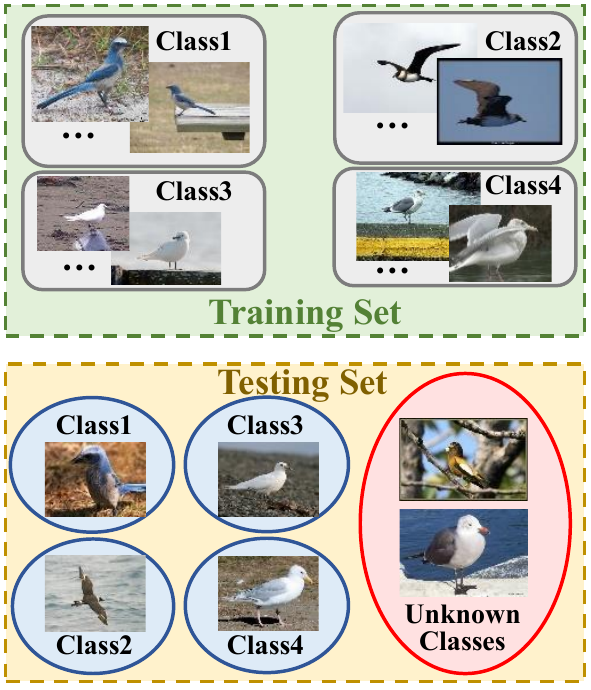}
	\end{center}
	\caption{Example on open-set fine-grained image recognition (OSFGR) from the public CUB dataset \cite{CUB}: The upper sub-figure shows four known classes (Florida jay, pomarine jaeger, ivory gull, and glaucous-winged gull) in the training set, the bottom sub-figure shows the four known classes and two unknown classes (evening grosbeak and heermann gull). An OSFGR model aims to simultaneously detect the unknown classes (the red circle) and classify the known classes (the blue circles).}
	\label{fig: concept}
\end{figure}

\IEEEPARstart{W}{ith} the rapid development of deep neural networks (DNNs), the computer vision community has witnessed significant progress in various image recognition tasks. Compared with generic image recognition, open-set fine-grained image recognition (OSFGR) which aims to simultaneously classify known-class fine-grained samples and detect unknown-class fine-grained samples, is a more challenging task due to the lower interclass discrepancies, as shown in Fig. \ref{fig: concept}.
Accordingly, this task requires OSFGR models to learn more discriminative features from input images \cite{IEEE-Access, good-closed-set}.

Recently, the spatial self-attention mechanism has been extensively used for feature learning in the computer vision community due to its ability of capturing local discriminative information, and particularly, the transformer and its variants \cite{ViT, TIT, PVT, T2T-ViT, Swin}, which were explored according to the spatial self-attention mechanism, have achieved a tremendous success in various visual tasks, such as image recognition
%\cite{transformer-recognition1,  transformer-recognition4},
\cite{transformer-recognition-TCSVT1, MoEP-AE},
%semantic segmentation
object tracking
%\cite{transformer-segmentation1,  transformer-segmentation4},
\cite{transformer-tracking-TCSVT1,  transformer-tracking-TCSVT2},
%object detection 
pose estimation
%\cite{transformer-object-detection1, transformer-object-detection4},
\cite{transformer-pose-TCSVT1, transformer-pose-TCSVT2}, and
monocular depth estimation \cite{transformer-monocular1, transformer-monocular2}. However, we empirically find that regardless of whether the scenario is open-set or closed-set, when a typical vision transformer is trained for object recognition with a set of object images that belong to different categories but have similar appearances, it sometimes could not pay accurate attention on objects, resulting in wrong predictions. This phenomenon would be described in detail in Sec. \ref{Deficiency}.

%\renewcommand{\thefootnote}{}
%\footnotetext{Code is available at \href{https://github.com/sjy-1995/STAN-OSFGR-code}{https://github.com/sjy-1995/STAN-OSFGR-code}}

To address this problem, we propose a spatial-temporal attention network to learn more accurate attention maps from input images for recognizing fine-grained objects, called STAN, inspired by the temporal attention mechanism in brains \cite{new_brain1, new_brain2}. The proposed STAN network learns the spatial-temporal attention maps with four modules: (1) a self-attention backbone module (the typical swin transformer \cite{Swin} is simply used as this module here) is used for extracting latent features by implementing a sequence of spatial self-attention operations; (2) a spatial feature self-organizing (SFSO) module is explored for facilitating the model training; (3) a spatial-temporal feature learning (STFL) module is explored for modeling temporal dependency in the sequence of latent features outputted from the SFSO module and aggregating these features progressively via an LSTM (Long Short-Term Memory network); (4) a context-aware (CA) module is explored for preserving/forgetting the long-term memory of the STFL module.
Furthermore, we present a STAN-based method for dealing with the OSFGR task, called STAN-OSFGR, where the proposed STAN is firstly used for learning spatial-temporal features from input object images and then a linear classifier is used to classify these features.

The main contributions of this paper are as follows: 

\begin{enumerate}
		
	\item[-] We empirically analyze the ability of a typical vision transformer with the spatial self-attention mechanism for discriminating fine-grained object images, finding that the vision transformer could not provide accurate attention maps sometimes. This analysis could contribute to a better understanding of the vision transformer.

	\item[-] We propose the spatial-temporal attention network STAN for learning both effective attentions and fine-grained feature representations. It is expected to be used as a feature extractor in various visual tasks, including but not limited to closed-set recognition and open-set recognition.
	
	\item[-] We design the STAN-OSFGR method for handling the OSFGR task with the proposed STAN. The priority of the designed STAN-OSFGR to nine state-of-the-art methods is demonstrated in Sec. \ref{Experiments}.

\end{enumerate}

The remainder of this paper is organized as follows. Sec.~\ref{Related_Work} reviews the related works. Sec.~\ref{Method} empirically investigates the ability of a typical vision transformer for discriminating fine-grained images, and describes the proposed STAN network and STAN-OSFGR method in detail. Sec.~\ref{Experiments} gives the experimental results. Sec.~\ref{Conclusion} concludes this paper.

\section{Related Works}    \label{Related_Work}

In this section, we review the related works on open-set recognition (OSR)/OSFGR and attention mechanisms, respectively.

\begin{figure*}
	\begin{center}
		\setlength{\abovecaptionskip}{0.cm}
		\includegraphics[height=12.8cm,width=19cm]{{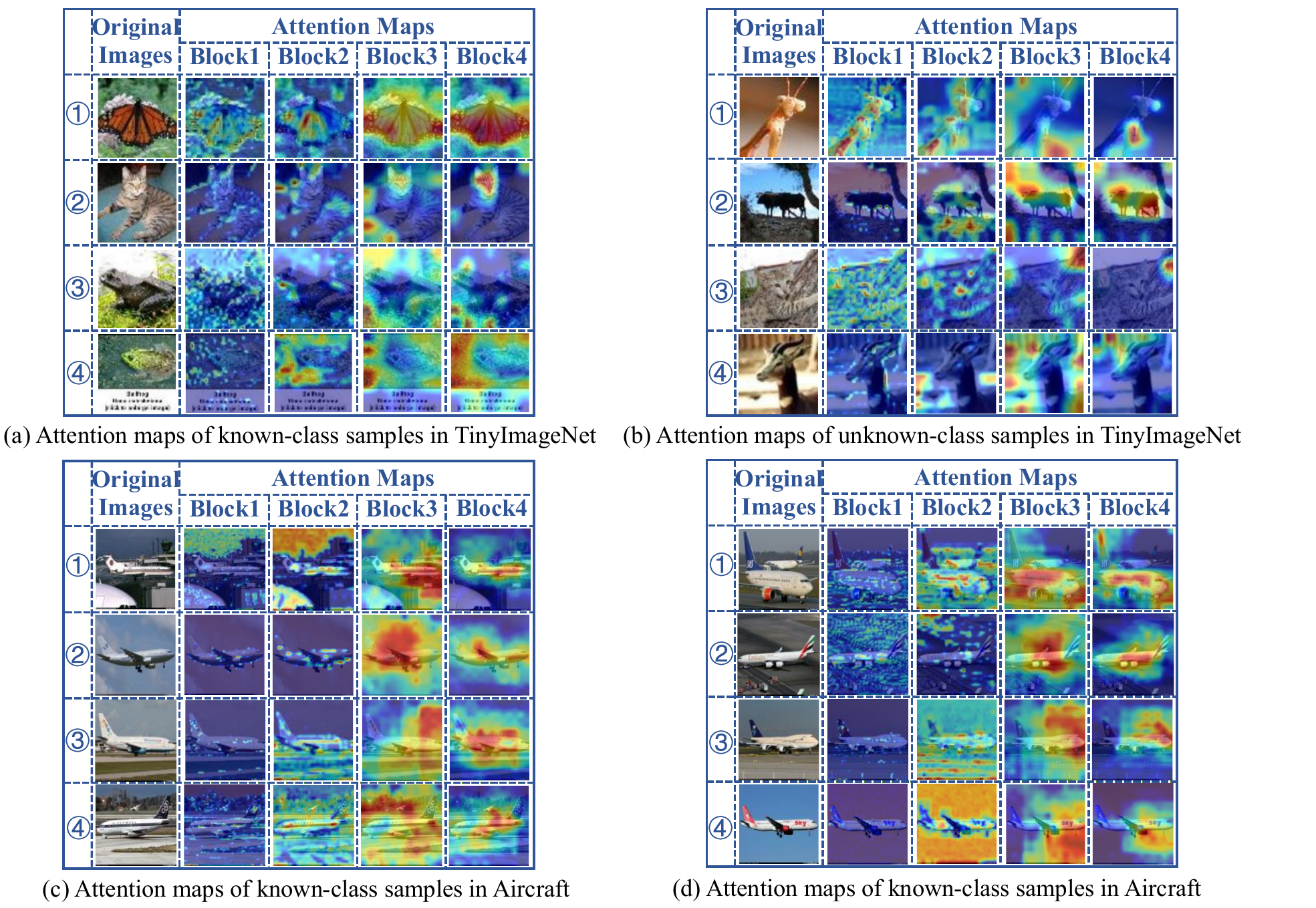}}
	\end{center}
	\caption{Visualization of the attention maps learnt by the swin transformer \cite{Swin} for some example images from the TinyImageNet dataset (sub-figures (a) and (b)), and the Aircraft dataset (sub-figures (c) and (d)). The sub-figures (a) and (c) show the examples of known classes, while the sub-figures (b) and (d) show the examples of unknown classes. The examples \normalsize{\textcircled{\scriptsize{1}}} and \normalsize{\textcircled{\scriptsize{2}}} in each sub-figure are predicted correctly by the swin transformer, while the examples \normalsize{\textcircled{\scriptsize{3}}} and \normalsize{\textcircled{\scriptsize{4}}} are predicted wrongly. The attentions in red regions are strongest, while the attentions in yellow, green, and blue regions decrease by degrees.}
%	It can be seen that the attentions focus on the objects more accurately at higher blocks/layers, and the transformer has already learnt accurate attentions on coarse-grained objects at the highest block/layer. However, the attentions learnt for the fine-grained objects are not accurate enough.}
	\label{fig: Swin-B backbone}
\end{figure*}

\subsection{OSR/OSFGR}

From a technical point of view, all the existing methods for handling the OSR task could be straightforwardly used in the OSFGR task without regard to their recognition accuracies. Most of the existing OSR methods in literature have been only evaluated on several coarse-grained datasets in their original papers, such as CIFAR+10/+50 \cite{CIFAR10, CIFAR100} and TinyImageNet \cite{TinyImageNet}. These methods generally learn discriminative feature representations either by utilizing only known-class samples
%\cite{OSR_deep_first, CAC, PROSER, CROSR, C2AE, GDFR, RPL, CPN, PMAL, P-ODN, OpenHybrid, CGDL, GMVAE-OSR, Capsule, MoEP-AE, OSR_transformer1, OSR_transformer2, OSR2022_3}
\cite{OSR_deep_first, PROSER, CROSR, C2AE, GDFR, RPL, CPN, PMAL, OpenHybrid, CGDL, GMVAE-OSR, Capsule, MoEP-AE, OSR_transformer2, OSR2022_3, GCPL, GMVAE-OSR, ARPL}
or by jointly utilizing both real known-class samples and synthesized unknown-class samples \cite{G-openmax, OSRCI, OpenGAN, OSR2022_1, OSR2022_2}. Zhang \emph{et al.} \cite{OpenHybrid} proposed to train a resflow network and a classifier in a latent feature space simultaneously for calculating both likelihood scores and classification scores. 
Yang \emph{et al.} \cite{GCPL} learnt the prototypes for modeling the distribution of each known class as a Gaussian mixture distribution, then the unknown classes could be detected according to whether the testing features belonged to these distributions. Cao \emph{et al.} \cite{GMVAE-OSR} modeled these distributions by using an autoencoder structure, GMVAE \cite{GMVAE}.
Sun \emph{et al.} \cite{MoEP-AE} proposed to encode the distributions of known-class sample features as multiple exponential power distributions for representing complex distributions that could not be modeled by Gaussian distributions. 
Kong and Ramanan \cite{OpenGAN} introduced some exposed outlier samples into the training set and used the discriminator of a GAN for detecting unknown classes. Chen \emph{et al.} \cite{ARPL} minimized the overlap of the distributions of known-class features and those of unknown-class features by adversarial learning using the reciprocal points.
%Cai \emph{et al.} \cite{OSR_transformer1} attached an additional detection head to the feature representations extracted from a vision transformer for sample clustering (Trans-ADH). 
Azizmalayeri and Rohban \cite{OSR_transformer2} revisited several training augmentations and chose the most suitable combination for the SoftMax-based OSR method using a vision transformer backbone.

Additionally, a few OSFGR methods \cite{IEEE-Access, good-closed-set} that aim at learning feature representations for recognizing fine-grained objects in open-set scenarios have been proposed and evaluated on a few fine-grained datasets, such as CUB \cite{CUB} and Aircraft \cite{Aircraft}. Dai \emph{et al.} \cite{IEEE-Access} used the class activation mapping values rather than the softmax scores for preserving information in the classification scores. Vaze \emph{et al.} \cite{good-closed-set} trained a closed-set classification network with the ResNet50 backbone using multiple training strategies for improving the generalization ability of the closed-set recognition model on open-set classes.

\subsection{Attention Mechanisms}

Existing attention mechanisms in the computer vision community could be divided into four categories according to their operating domains \cite{attention1}: channel attention, spatial attention, temporal attention, and branch attention. Among them, the spatial attention is more widely used. As an essential branch of spatial attention, the spatial self-attention mechanism, which captures the internal correlation of the features, has experienced from stand-alone modules \cite{attention2, attention3, attention4, attention5, attention6} to integrated networks \cite{ViT, TIT, PVT, T2T-ViT, Swin}. Examples of the latter include the vision transformer and its variants \cite{ViT, TIT, PVT, T2T-ViT, Swin}, which integrate a sequence of spatial self-attention operations in their networks. In \cite{ViT}, the vision transformer was proposed to rely entirely on self-attention in the form of fully-connected layers along with position encoding for handling several image tasks, which was the pioneering work to introduce transformers into the computer vision community and had the advantage in learning global information. Its variants utilized embedded sub-transformer \cite{TIT}, pyramid architecture \cite{PVT}, layer-wise tokens-to-tokens transformation \cite{T2T-ViT}, or shifted windows \cite{Swin} for excavating local information that was insufficiently learnt by the vanilla vision transformer to improve the model performance.

However, to our best knowledge, there is no report on whether accurate attention maps can be obtained from the fine-grained object images by a vision transformer. In this work, we investigate the ability of a typical vision transformer for discriminating fine-grained object images, and propose a spatial-temporal attention network for both learning more accurate attention maps and improving the performance of open-set fine-grained recognition, which would be described in the following section.

\begin{figure*}[t]
	\begin{center}
		\setlength{\abovecaptionskip}{0.cm}
		\includegraphics[height=7cm,width=16.9cm]{{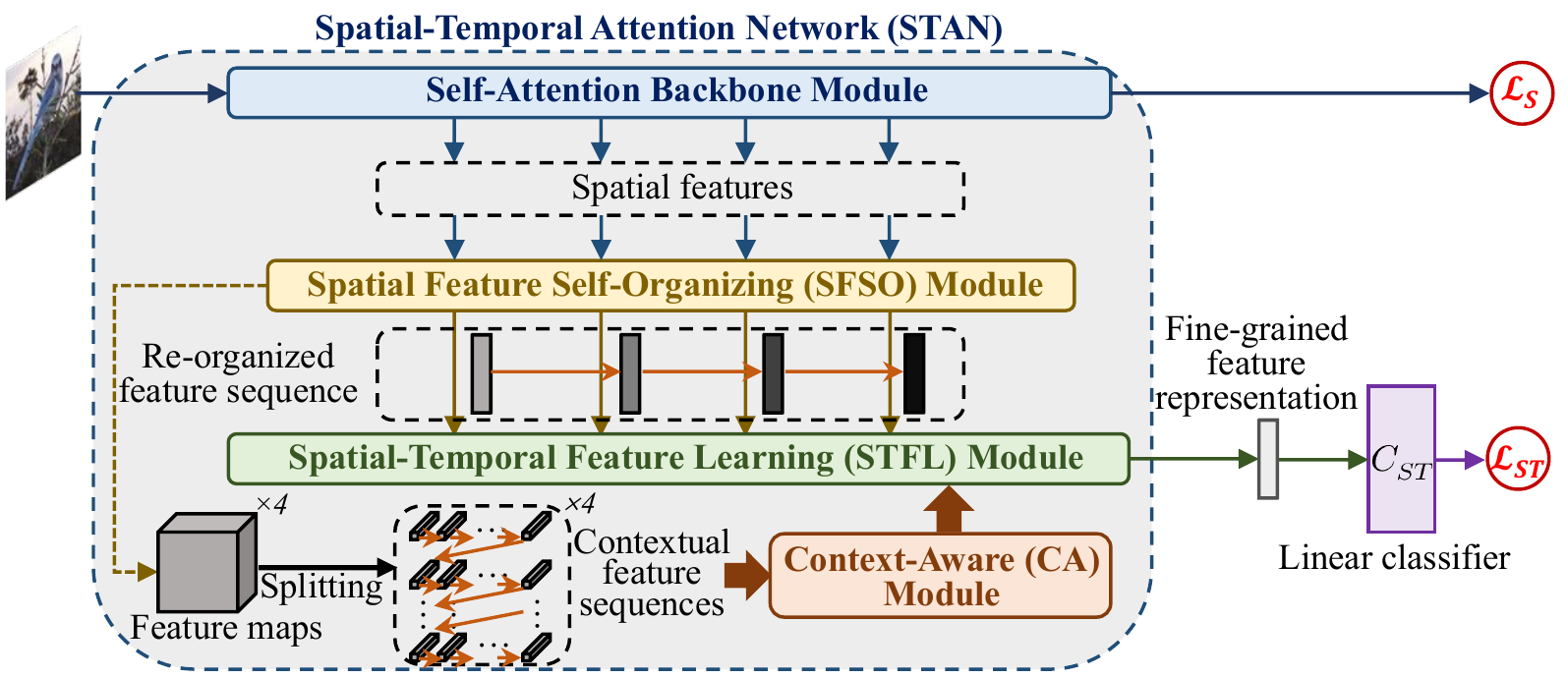}}
	\end{center}
	\caption{Architecture of the proposed spatial-temporal attention network (STAN): STAN consists of four modules, including a self-attention backbone module for obtaining a sequence of spatial features, a spatial feature self-organizing (SFSO) module for facilitating the model training, a spatial-temporal feature learning (STFL) module for learning fine-grained feature representation by modeling temporal dependency via an LSTM, and a context-aware (CA) module for preserving/forgetting the long-term memory via contextual information. STAN is trained by minimizing a weighted sum of two loss terms, including a spatial-feature classification loss term $\mathcal{L}_{S}$ for learning a sequence of features under the spatial attention mechanism, and a spatial-temporal-feature classification loss term $\mathcal{L}_{ST}$ for learning fine-grained spatial-temporal feature representations.}
	\label{fig: STAN}
\end{figure*}

\section{STAN for OSFGR}  \label{Method}
In this section, we firstly investigate the ability of a typical transformer in both coarse-grained and fine-grained image recognition tasks regardless of open-set or closed-set scenarios. Then we describe the proposed spatial-temporal attention network STAN in detail. Finally, we introduce the proposed STAN-OSFGR method based on the STAN network for handling the open-set fine-grained task.

\subsection{Could a transformer pay accurate attention on objects in images?}    \label{Deficiency}

Vision transformers \cite{ViT, TIT, PVT, T2T-ViT, Swin} have shown their effectiveness in object recognition, mainly due to their multiple spatial self-attention operations. Then, a question is naturally raised: \textit{`Could a transformer learn accurate attention maps on objects in images?'} To address this question, the swin transformer \cite{Swin}, which is a typical vision transformer, is trained with the traditional cross-entropy classification loss for object classification by utilizing a portion of classes from the TinyImageNet dataset \cite{TinyImageNet} (a coarse-grained dataset) and the Aircraft dataset \cite{Aircraft} (a fine-grained dataset) respectively.
%a typical vision transformer (\emph{i.e.}, swin transformer \cite{Swin} is trained with the traditional cross-entropy classification loss and tested by a threshold-based strategy for handling the OSR task on both a coarse-grained dataset (\emph{i.e.}, the TinyImageNet dataset \cite{TinyImageNet}) and a fine-grained dataset (\emph{i.e.}, the Aircraft dataset \cite{Aircraft}). 
%Then, we take four known-class images and four unknown-class images from each dataset as examples which are predicted correctly or wrongly by the transformer, and the corresponding attention maps learnt by the transformer at the four blocks are visualized by Grad-CAM \cite{Grad-CAM}, which are shown in Fig. \ref{fig: Swin-B backbone}. 
For each dataset, we use Grad-CAM \cite{Grad-CAM} to visualize the attention maps of 8 example images from the corresponding testing set (4 images from the known classes which are observed in the training set, while the rest from the unknown classes which are not observed in the training set), outputted from the four blocks of the swin transformer as shown in Fig. \ref{fig: Swin-B backbone}.

As seen from the examples \normalsize{\textcircled{\scriptsize{1}}} and \normalsize{\textcircled{\scriptsize{2}}} (which are predicted correctly) in Fig. \ref{fig: Swin-B backbone} (a) and Fig. \ref{fig: Swin-B backbone} (b), 
%the attention tends to concentrate as the block/layer deepens, and a good attention map at the highest block (\emph{i.e.}, the Block 4) is often accompanied by a good prediction. 
a higher block of the transformer is prone to pay more attention on the objects in the example images, resulting in a correct prediction. However, as seen from the examples \normalsize{\textcircled{\scriptsize{3}}} and \normalsize{\textcircled{\scriptsize{4}}} (which are predicted wrongly) in Fig. \ref{fig: Swin-B backbone} (a) and Fig. \ref{fig: Swin-B backbone} (b), the attention maps learnt by the higher blocks of the transformer are not accurate, which focus on the backgrounds (the grasses and water) or the distractors (the trunk and fences), resulting in wrong predictions. 
%For known-class objects such as the monarch butterfly and the tabby cat, the final attentions learnt by the transformer concentrate well on the discriminative parts of the objects such as wings and the face. For unknown-class objects such as mantis and ox, the final attentions can also focus on the discriminative regions which effectively distinguish these objects from the known-class ones. In the above examples, the dragonfly compound eye and the bull horn focused on by the transformer are almost non-existent in the known-class images. However, as seen from the examples \normalsize{\textcircled{\scriptsize{3}}} and \normalsize{\textcircled{\scriptsize{4}}} in the two sub-figures, the two bullfrog images are wrongly predicted as `tailed frog' since the transformer focuses on the background regions (grasses and the water) that are shared between bullfrog images and tailed frog images for classification, and the Egyptian cat image as well as the gazelle image are wrongly predicted as known-class objects (`birdhouse' and `viaduct', respectively) since the attention maps highlight the distractors (the trunk and the fence) whose texture or shape is similar to that of the clues in birdhouse images and viaduct images. 

The issue that the transformer focuses on the backgrounds or distractors more easily for classification is not common in the coarse-grained OSR task, because images in different classes usually have significantly different appearances. However, it is a common issue in the OSFGR task where fine-grained objects usually appear in similar environments. As seen from the two sub-figures Fig. \ref{fig: Swin-B backbone} (c) and Fig. \ref{fig: Swin-B backbone} (d), though the images in the examples \normalsize{\textcircled{\scriptsize{1}}} and \normalsize{\textcircled{\scriptsize{2}}} are correctly predicted since the attention maps learnt by the higher blocks of the transformer mainly highlight the aircrafts, these attentions are universally not accurate enough, which focus on the sky or runways to different extent.

%As seen from the two sub-figures Fig. \ref{fig: Swin-B backbone} (c) and Fig. \ref{fig: Swin-B backbone} (d), an accurate attention map at the Block 4 plays a more important role for resulting in a correct prediction: The attentions in examples \normalsize{\textcircled{\scriptsize{1}}} and \normalsize{\textcircled{\scriptsize{2}}} mainly focus on the aircraft objects as well as slightly highlight the backgrounds and distractors (sky and runways), which results in correct predictions; The attention maps in the examples \normalsize{\textcircled{\scriptsize{3}}} and \normalsize{\textcircled{\scriptsize{4}}} show great deviations between the aircraft objects and the learnt attentions which focus on the sky or runways rather than the discriminative parts of the aircrafts, and these attention maps result in wrong predictions. 

In sum, it is revealed to some extent from the above experimental results that the transformer could sometimes not learn accurate attention maps for coarse-grained object images, and they universally could not learn attention maps for fine-grained object images which are accurate enough even some of these images could be accurately predicted.

\begin{figure}[t]
	\begin{center}
		\setlength{\abovecaptionskip}{0.cm}
		\includegraphics[height=2.7cm,width=8.8cm]{{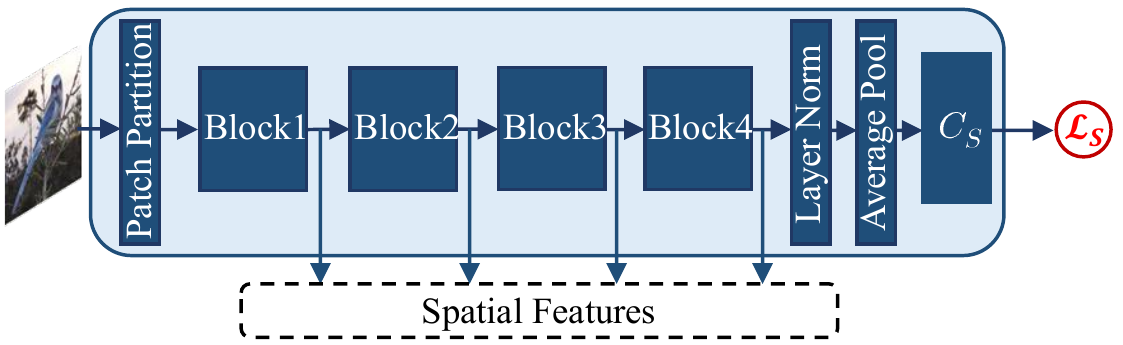}}
	\end{center}
	\caption{Architecture of swin transformer \cite{Swin}. The input image is firstly split into multiple patches and passed through a linear embedding layer to obtain latent features. Then, these features are passed through four integrated blocks with spatial self-attention operations. Hierarchical spatial features can be produced by the four blocks. Next, a normalization layer and a pooling layer are performed. Then the feature is inputted to a linear classifier $C_S$.}
	\label{fig: module0}
\end{figure}

\subsection{Spatial-Temporal Attention Network}  \label{STAN-modules}

To address the above revealed issue, here, we propose a spatial-temporal attention network (called STAN) for learning fine-grained feature representations, whose architecture is shown in Fig. \ref{fig: STAN}. STAN consists of four modules: a self-attention backbone module for learning sequential features with multiple spatial self-attention operations from input images, a spatial feature self-organizing (SFSO) module for transforming the learnt sequential features to facilitate the model training, a spatial-temporal feature learning (STFL) module for learning fine-grained feature representation via an LSTM, and a context-aware (CA) module that is embedded into the STFL module for selectively preserving the long-term memory. In this work, we simply use swin transformer \cite{Swin}, which contains 4 spatial self-attention blocks and a spatial-feature classification loss as shown in Fig. \ref{fig: module0}, as the self-attention backbone module. Accordingly, a sequence of features could be obtained from the four spatial self-attention blocks of the swin transformer, and they are inputted to the SFSO module. In the following parts, the SFSO, STFL, and CA modules would be described in detail.

\begin{figure}[t]
	\begin{center}
		\setlength{\abovecaptionskip}{0.cm}
		\includegraphics[height=7cm,width=7.2cm]{{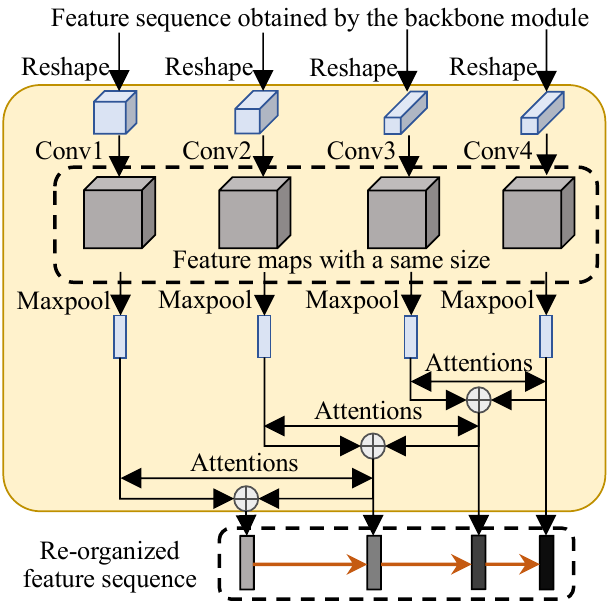}}
	\end{center}
	\caption{Architecture of the spatial feature self-organizing (SFSO) module. The feature sequence is reshaped and passed into convolutional layers for obtaining the sequence of feature maps of a same size. Next, these feature maps are passed through max-pooling layers for dimensionality reduction. Then, the features from the higher blocks are aggregated with the features from the lower blocks with self-attentions.}
	\label{fig: module1}
\end{figure}

\begin{figure}[t]
	\begin{center}
		\setlength{\abovecaptionskip}{0.cm}
		\includegraphics[height=9.8cm,width=8.5cm]{{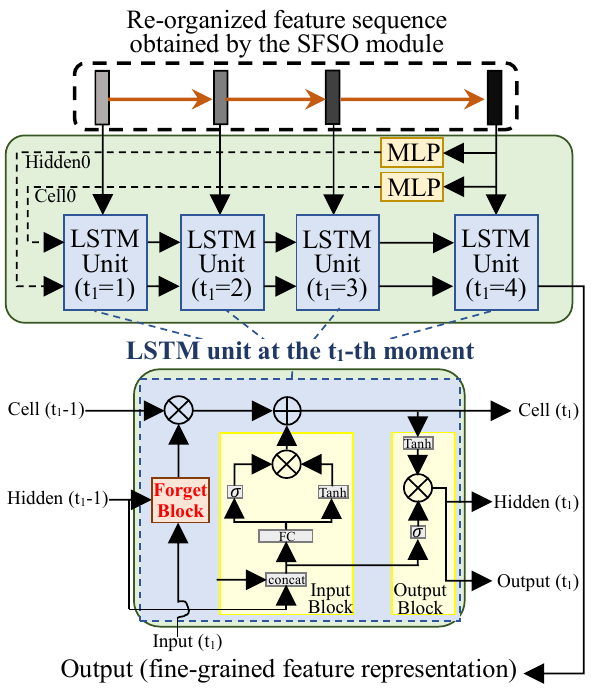}}
	\end{center}
	\caption{Architecture of the spatial-temporal feature learning (STFL) module: The upper sub-figure shows the holistic architecture of this module which mainly consists of a single-layer LSTM with 4 moments, and the bottom sub-figure shows the internal architecture of the LSTM unit which consists of a memory cell and three blocks: an input block, an output block, and a forget block by convention. The symbols `$\otimes$' and `$\oplus$' represent element-wise dot product and addition, respectively. `$\sigma$', `tanh', and `FC' represent a Sigmoid layer, a Tanh layer, and a fully-connected layer, respectively. The operation `concat' performs concatenation between two features.}
	\label{fig: module2}
\end{figure}

\subsubsection{Spatial Feature Self-Organizing (SFSO) Module}

It is noted that (i) the sizes of the features outputted from the blocks of the self-attention backbone module might be different (also for the swin transformer \cite{Swin}); (ii) for most of DNNs in literature \cite{Swin, DNNs1, DNNs2, DNNs3}, the features learnt from their lower blocks/layers generally have a lower discriminability than those from their higher blocks/layers. Hence here, the SFSO module is explored to take the sequence of features outputted from the self-attention backbone module as its input, and then re-organize them into a sequence of features so that (i) all the re-organized features are of a same size and (ii) each re-organized feature could integrate the original features outputted from higher blocks. Fig. \ref{fig: module1} shows the architecture of the designed SFSO module. 

As seen from Fig. \ref{fig: module1}, the sequence of features of different sizes outputted from the backbone module is firstly reshaped respectively, and then convolution (2-d convolution) operations are implemented on these feature maps in the SFSO module, so that this sequence of features of different sizes is transformed into a sequence of features of a same size. Next, the obtained features of a same size are passed through max-pooling layers respectively and 
%are further injected into a feature pyramid structure
further injected into a high-to-low feature aggregation structure with self-attentions, and a sequence of re-organized features where each feature integrates the original features outputted from the higher blocks of the backbone module is obtained.

\begin{figure}[t]
	\begin{center}
		\setlength{\abovecaptionskip}{0.cm}
		\includegraphics[height=9cm,width=6.8cm]{{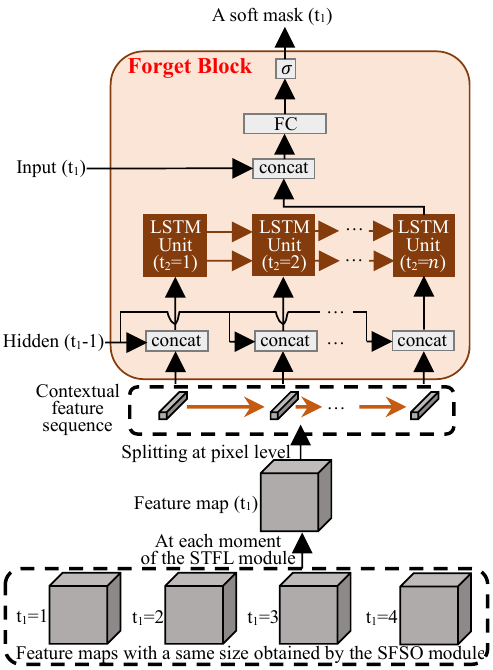}}
	\end{center}
	\caption{Architecture of the context-aware (CA) module. This module is implemented as the forget block of the STFL module via another single-layer LSTM with a conventional structure. Among the three inputs, the contextual feature sequence is obtained by splitting each feature map in the feature map sequence obtained by the SFSO module at the pixel level. Then, the contextual features are concatenated with the Hidden state and inputted to the LSTM. The output of the LSTM at the final moment is concatenated with the input feature and passed through a fully-connected layer and a sigmoid layer for obtaining a soft mask.}
	\label{fig: module3}
\end{figure}

\subsubsection{Spatial-Temporal Feature Learning (STFL) Module}

The STFL module is explored to facilitate the model training. It takes the features re-organized by the SFSO module as its input, and outputs an aggregated feature that has temporal dependency on the re-organized features as the fine-grained feature representation. The architecture of this module is shown in Fig. \ref{fig: module2}.

As seen from Fig. \ref{fig: module2}, a single-layer LSTM is adopted here, which accepts the re-organized feature sequence that consists of 4 moments of features. The LSTM unit which is unrolled/unfolded through moments comprises of a (long-term) memory cell (for retaining global memory of the LSTM over all of the 4 moments in 4 Cell states) and 3 blocks (the input/output/forget block) for determining what information to be added/exported/dropped from the memory. The output of the previous moment is utilized as the Hidden state, while the output of the final moment is used as the fine-grained feature representations to be classified. Besides, the initial Cell and Hidden states are instance-adaptive vectors that are mapped from the input feature at the final moment for accelerating the model training.

\subsubsection{Context-Aware (CA) Module}

The CA module is explored to preserve/forget information selectively from the memory cell. It is implemented as the forget block of the STFL module. It contains three inputs: the Hidden state at the previous moment in the STFL module, the input feature at the current moment in the STFL module, and sequential contextual features obtained by splitting the current feature map at the pixel level. It outputs a soft mask, whose elements are limited to $(0,1)$, for masking the Cell state of the STFL module. The architecture of this module is shown in Fig. \ref{fig: module3}.

As seen from Fig. \ref{fig: module3}, another single-layer LSTM with $n$ moments (where $n$ is the number of pixels in the feature map) is used here for converting the three inputs into the final soft mask. The structure of this LSTM unit adopts a standard form, whose initial Cell and Hidden states are randomized. The input at the current moment of this LSTM is a sequence of features where each split feature is concatenated with the Hidden state of the STFL module, and the output of this LSTM is concatenated with the input feature of the STFL module. The concatenated feature is passed through a fully-connected layer and a sigmoid layer for obtaining the final soft mask.

\subsection{STAN-OSFGR}

In this subsection, we introduce the STAN-OSFGR method for handling the OSFGR task, which integrates the proposed STAN with a linear classifier (\emph{i.e.}, the classifier $C_{ST}$ shown in Fig. \ref{fig: STAN}). The STAN-OSFGR method employs a joint classification loss function for training and a threshold-based strategy for inference as follows:

\textbf{Loss.} The used joint classification loss function contains two loss terms, including a spatial-feature classification loss term $\mathcal{L}_{S}$ and a spatial-temporal-feature classification loss term $\mathcal{L}_{ST}$, as shown in Fig. \ref{fig: STAN}:

\begin{enumerate}
	
	\item[-] $\mathcal{L}_{S}$: The spatial-feature classification loss term is used for measuring the difference between the predictions of the classifier $C_S$ in the self-attention backbone module and the groundtruths. It encourages the backbone module to learn a sequence of features under the spatial self-attention mechanism. Here, $\mathcal{L}_{S}$ employs the traditional cross-entropy loss as:
	\begin{align}
	\mathcal{L}_{S} = - \frac{1}{N} \sum_{i=1}^{N} \mathrm{log}p_{i}
	\end{align}
	where $N$ is the batch size, and $p_{i}$ is the classification probability obtained by implementing the \emph{SoftMax} layer for the classification scores outputted from the classifier $C_{S}$.
	
	\item[-] $\mathcal{L}_{ST}$: The spatial-temporal-feature classification loss term is used for measuring the difference between the predictions of the introduced classifier $C_{ST}$ and the groundtruths. It encourages the STFL module to learn fine-grained spatial-temporal feature representations. Here, $\mathcal{L}_{ST}$ also employs the traditional cross-entropy loss as:
	\begin{align}
	\mathcal{L}_{ST} = - \frac{1}{N} \sum_{j=1}^{N} \mathrm{log}q_{j}
	\end{align}
	where $q_{j}$ is the corresponding classification probability calculated from the classification scores which are outputted from the classifier $C_{ST}$.

\end{enumerate}

\begin{table*}[ht]
	\centering
	\caption{Numbers of known/unknown classes in different settings/configurations/datasets. `Easy', `Medium', and `Hard' are three difficulty modes in the CUB and Aircraft benchmark datasets.}
	\begin{tabular}{m{2.2cm}<{\raggedright}m{2.5cm}<{\centering}m{2.2cm}<{\centering}m{1.5cm}<{\centering}m{1.5cm}<{\centering}m{1.5cm}<{\centering}m{1.5cm}<{\centering}}
		\toprule
		Setting & Configuration & Dataset & Known & \multicolumn{3}{c}{Unknown}  \\
		\midrule
		\multirow{7}{*}[-9pt]{fine-grained} & \multirow{5}{*}[-9pt]{standard-dataset} & \multirow{2}{*}[-3pt]{CUB} & \multirow{2}{*}[-3pt]{100} & Easy & Medium & Hard \\
		\cmidrule{5-7}
		& & & & 32 & 34 & 34 \\
		%		\midrule
		\cmidrule{3-7}
		& & \multirow{2}{*}[-3pt]{Aircraft} & \multirow{2}{*}[-3pt]{50} & Easy & Medium & Hard \\
		\cmidrule{5-7}
		& & & & 20 & 17 & 13 \\
		%		\midrule
		\cmidrule{3-7}
		& & Stanford-Cars & 98 & \multicolumn{3}{c}{98} \\
		\cmidrule{2-7}
		& \multirow{2}{*}[-1pt]{cross-dataset} & CUB & 50 & \multicolumn{3}{c}{200} \\
		& & Stanford-Cars & 50 & \multicolumn{3}{c}{196} \\	
		\midrule
		\multirow{2}{*}[-3pt]{coarse-grained} & \multirow{2}{*}[-3pt]{--} & CIFAR+10/+50 & 10 & \multicolumn{3}{c}{10/50}\\
		%		\midrule
		\cmidrule{3-7}
		& & TinyImageNet & 20 & \multicolumn{3}{c}{180}\\
		\bottomrule
	\end{tabular}
	\label{table: num_classes}
\end{table*}

Hence, the total joint classification loss function is the weighted sum of $\mathcal{L}_{S}$ and $\mathcal{L}_{ST}$:
\begin{align}
\mathcal{L} = \mathcal{L}_{S} + \lambda \mathcal{L}_{ST}
\end{align}
where $\lambda$ is a hyperparameter for balancing the two loss terms.

\textbf{Inference.} The classification scores outputted from the classifier $C_{ST}$ are used for inference. Specifically, the maximum value of the logit vector obtained from a testing image is used as a score for detecting unknown classes as well as classifying known classes:
\begin{align}
\mathrm{score} = \max \limits_{k} \{ f_k \}
\end{align}
where $f_k$ is the $k$-th ($k=1,2,...,K$) value of the logit vector outputted from the classifier $C_{ST}$, and $K$ is the number of known classes.

Then, if the score is larger than a threshold $\theta$, the testing image will be recognized as one of the known classes; otherwise, it will be recognized as unknown classes. Besides, in the former case, the argmax index in the logit vector indicates the predicted label:
\begin{align}
\mathrm{prediction} =
\begin{cases}
\mathrm{argmax}_{k \in \{1,2,...,K \}} f_k & \text{if $\mathrm{score} > \theta$} \\
\mathrm{unknown \ classes}& \text{if $\mathrm{score} \leq \theta$}
\end{cases}
\end{align}

\section{Experiments}  \label{Experiments}
In this section, we firstly introduce the benchmark datasets and the evaluation metrics. Then, we introduce the implementation details. Next, we conduct comparative experiments for evaluating the proposed method. Then, we analyze the hyperparameter in the total loss. Next, we conduct two ablation studies. Lastly, we end up with visualization.

\subsection{Benchmark Datasets and Evaluation Metrics}

\subsubsection{Benchmark Datasets}

In order to comprehensively evaluate the proposed STAN-OSFGR method, we conduct experiments under both the fine-grained and coarse-grained settings as follows:

For the fine-grained setting, our method is evaluated under two configurations: the standard-dataset configuration where the known-class images and unknown-class images are taken from the same dataset and the cross-dataset configuration where the known-class images and unknown-class images are taken from different datasets. The experiments under both the configurations are conducted based on three public datasets, including the CUB dataset \cite{CUB}, the Aircraft dataset \cite{Aircraft}, and the Stanford-Cars dataset \cite{Stanford-Cars}. The numbers of the split known/unknown classes in these datasets under the two configurations are listed in Table \ref{table: num_classes}. The standard-dataset configuration is deployed based on these datasets in the following manner:

%\begin{table}[t]
%	\centering
%	\caption{Numbers of known/unknown classes in different settings/configurations/datasets. `Easy', `Medium', and `Hard' are three difficulty modes in the CUB and Aircraft benchmark datasets.}
%	\begin{tabular}{m{2.2cm}<{\raggedright}m{1cm}<{\centering}m{1cm}<{\centering}m{1cm}<{\centering}m{1cm}<{\centering}}
%		\toprule
%		Dataset & Known & \multicolumn{3}{c}{Unknown}  \\
%		\midrule
%		\multirow{2}{*}[-3pt]{CUB} & \multirow{2}{*}[-3pt]{100} & Easy & Medium & Hard \\
%		\cmidrule{3-5}
%		& & 32 & 34 & 34 \\
%		\midrule
%		\multirow{2}{*}[-3pt]{Aircraft} & \multirow{2}{*}[-3pt]{50} & Easy & Medium & Hard \\
%		\cmidrule{3-5}
%		& & 20 & 17 & 13 \\
%		\midrule
%		Stanford-Cars & 98 & \multicolumn{3}{c}{98}\\
%		\midrule
%		CIFAR+10/+50 & 10 & \multicolumn{3}{c}{10/50}\\
%		\midrule
%		TinyImageNet & 20 & \multicolumn{3}{c}{180}\\
%		\bottomrule
%	\end{tabular}
%	\label{table: num_classes}
%\end{table}

\begin{table*}
	\centering
	\caption{Comparison of OSFGR results on the Aircraft dataset under the standard-dataset configuration.}
	\begin{tabular}{m{4.2cm}<{\raggedright}m{1cm}<{\centering}m{4.2cm}<{\centering}m{4.2cm}<{\centering}}
		\toprule
		Method & ACC & \makecell[c]{AUROC\\ (Easy/Medium/Hard)} & \makecell[c]{OSCR\\ (Easy/Medium/Hard)} \\
		\midrule
		OpenHybrid \cite{OpenHybrid} & 0.735 & 0.884/0.844/0.780 & 0.718/0.687/0.629  \\
		OpenGAN \cite{OpenGAN} & 0.807 & 0.841/0.821/0.688 & 0.795/0.754/0.651  \\
		ARPL \cite{ARPL} & 0.917 & 0.867/0.852/0.674 & 0.814/0.820/0.657  \\
		GCPL \cite{GCPL} & 0.823 & 0.837/0.828/0.709 & 0.805/0.761/0.652   \\
		GMVAE-OSR \cite{GMVAE-OSR} & 0.814 & 0.849/0.833/0.687 & 0.799/0.753/0.661    \\
		CAMV \cite{IEEE-Access} & 0.868 & 0.880/0.844/0.730 & 0.801/0.772/0.680  \\
		Cross-Entropy+ \cite{good-closed-set} & 0.917 & 0.907/0.864/0.776 & 0.868/0.831/0.754  \\
		\midrule
		Backbone & 0.896 & 0.848/0.827/0.723 & 0.799/0.778/0.686  \\
		%		STAN-Transformer & 0.911 & 0.924/0.868/0.752 & 0.871/0.820/0.719 \\
		%		Trans-ADH \cite{OSR_transformer1} & 0.902 & 0.862/0.839/0.746 & 0.811/0.785/0.692   \\
		OpenHybrid(Swin) & 0.901 & 0.889/0.830/0.733 & 0.823/0.781/0.689  \\
		ARPL(Swin) & 0.915 & 0.880/0.833/0.720 & 0.836/0.793/0.691   \\
		Cross-Entropy+(Swin) & 0.903 & 0.859/0.841/0.712 & 0.814/0.796/0.681  \\
		Trans-AUG \cite{OSR_transformer2} & 0.898 & 0.873/0.842/0.739 & 0.814/0.787/0.689   \\
		MoEP-AE-OSR \cite{MoEP-AE} & 0.894 & 0.907/0.876/0.752 & 0.831/0.805/0.701  \\
		STAN-OSFGR & \textbf{0.927} & \textbf{0.932/0.897/0.840} & \textbf{0.888/0.858/0.807}  \\
		\bottomrule
	\end{tabular}
	\label{table: Aircraft}
\end{table*}

\begin{table*}
	\centering
	\caption{Comparison of OSFGR results on the CUB dataset under the standard-dataset configuration.}
	\begin{tabular}{m{4.2cm}<{\raggedright}m{1cm}<{\centering}m{4.2cm}<{\centering}m{4.2cm}<{\centering}}
		\toprule
		Method & ACC & \makecell[c]{AUROC\\ (Easy/Medium/Hard)} & \makecell[c]{OSCR\\ (Easy/Medium/Hard)} \\
		\midrule
		OpenHybrid \cite{OpenHybrid} & 0.683 & 0.873/0.862/0.739 & 0.662/0.649/0.531  \\
		OpenGAN \cite{OpenGAN} & 0.799 & 0.801/0.765/0.707 & 0.725/0.706/0.648  \\		
		ARPL \cite{ARPL} & 0.863 & 0.814/0.772/0.703 & 0.747/0.710/0.659  \\
		GCPL \cite{GCPL} & 0.783 & 0.805/0.732/0.645 & 0.711/0.619/0.565   \\
		GMVAE-OSR \cite{GMVAE-OSR} & 0.725 & 0.826/0.750/0.704 & 0.694/0.628/0.553   \\
		CAMV \cite{IEEE-Access} & 0.836 & 0.845/0.802/0.709 & 0.746/0.715/0.640  \\
		Cross-Entropy+ \cite{good-closed-set} & 0.862 & 0.883/0.823/0.763 & 0.798/0.754/0.708  \\
		\midrule
		Backbone & 0.949 & 0.945/0.875/0.804 & 0.908/0.848/0.781 \\
		%		STAN-Transformer & 0.947 & 0.962/0.908/0.822 & 0.921/0.875/0.798  \\
		%		Trans-ADH \cite{OSR_transformer1} & 0.953 & 0.950/0.881/0.811 & 0.912/0.851/0.796   \\
		OpenHybrid(Swin) & 0.950 & 0.953/0.881/0.808 & 0.918/0.855/0.783   \\
		ARPL(Swin) & 0.952 & 0.948/0.877/0.810 & 0.912/0.850/0.788   \\
		Cross-Entropy+(Swin) & 0.953 & 0.950/0.879/0.815 & 0.917/0.854/0.794    \\
		Trans-AUG \cite{OSR_transformer2} & 0.950 & 0.953/0.882/0.818 & 0.914/0.854/0.795   \\
		MoEP-AE-OSR \cite{MoEP-AE} & 0.948 & 0.957/0.889/0.814 & 0.915/0.856/0.787  \\
		STAN-OSFGR & \textbf{0.959} & \textbf{0.965/0.922/0.860} & \textbf{0.934/0.898/0.841}  \\
		\bottomrule
	\end{tabular}
	\label{table: CUB}
\end{table*}

\begin{table*}[t]
	\centering
	\caption{Comparison of OSFGR results on the Stanford-Cars dataset under the standard-dataset configuration.}
	\begin{tabular}{m{4.2cm}<{\raggedright}m{1cm}<{\centering}m{4.2cm}<{\centering}m{4.2cm}<{\centering}}
		\toprule
		Method & ACC & AUROC & OSCR  \\
		\midrule
		OpenHybrid \cite{OpenHybrid} & 0.675 & 0.699 & 0.637 \\
		OpenGAN \cite{OpenGAN} & 0.704 & 0.763 & 0.682 \\	
		ARPL \cite{ARPL} & 0.752 & 0.839 & 0.726 \\
		GCPL \cite{GCPL} & 0.688 & 0.735 & 0.663   \\
		GMVAE-OSR \cite{GMVAE-OSR} & 0.683 & 0.756 & 0.670   \\
		CAMV \cite{IEEE-Access} & 0.754 & 0.832 & 0.724 \\	
		Cross-Entropy+ \cite{good-closed-set} & 0.768 & 0.859 & 0.735 \\
		\midrule
		Backbone & 0.856 & 0.905 & 0.800 \\
		%		STAN-Transformer & 0.888 & 0.934  \\
		%		Trans-ADH \cite{OSR_transformer1} & 0.868 & 0.914 & 0.811   \\
		OpenHybrid(Swin) & 0.857 & 0.910 & 0.805  \\
		ARPL(Swin) & 0.860 & 0.908 & 0.807  \\
		Cross-Entropy+(Swin) & 0.864 & 0.906 & 0.808  \\
		Trans-AUG \cite{OSR_transformer2} & 0.860 & 0.910 & 0.809   \\
		MoEP-AE-OSR \cite{MoEP-AE} & 0.869 & 0.922 & 0.819  \\
		STAN-OSFGR & \textbf{0.888} & \textbf{0.936} & \textbf{0.849} \\
		\bottomrule
	\end{tabular}
	\label{table: Stanford-Cars}
	%	\\[-12pt]
\end{table*}

\begin{enumerate}

	\item[-] \textbf{CUB:} The Caltech-UCSD Birds (CUB) dataset \cite{CUB} contains 200 categories of bird images with label and attribute tags. Here, we use CUB-200-2011 version of this dataset as done in the state-of-the-art work \cite{good-closed-set} for handling the OSFGR task: 100 classes are selected as the known classes, while the rest as the unknown classes. Three difficulty modes are set for the unknown classes based on the attribute similarity between each unknown class and the known classes. Generally, the unknown classes in the `Easy' mode are more unlike the known classes thus can be relatively easily distinguished from the known classes, the unknown classes in the `Hard' mode are more similar to the known classes, thus making it more difficult for detecting semantic novelty, and the similarity and detection difficulty of the `Medium' mode is between the `Easy' mode and the `Hard' mode. We use the same data splits with that in \cite{good-closed-set}. 
	
	\item[-] \textbf{Aircraft:} The FGVC-Aircraft-2013b (Aircraft) dataset \cite{Aircraft} is another commonly used benchmark dataset in the fine-grained image classification task, and is also with both label and attribute tags. Its labels contain three levels: the manufacturer level with 30 categories, the family level with 70 categories, and the variant level with 100 categories. We choose the level of variants that consists of 100 categories of aircraft images, 50 classes of which are selected as the known classes, and the rest classes are further split into the three difficulty modes of unknown classes. We use the same data splits as done in \cite{good-closed-set}.
	
	\item[-] \textbf{Stanford-Cars:} The Stanford-Cars dataset \cite{Stanford-Cars} contains 196 categories of car images, whose labels depend on Make, Model, and Year. 
%	To our best knowledge, this dataset is more challenging and has not been used for evaluation in the OSFGR task in literature, and hence we design the following splitting manner for further evaluating the proposed method on this dataset: 
	we design the following splitting manner for further evaluating the proposed method on this dataset: the first 98 classes are selected as the known classes, and the rest 98 classes are used as the unknown classes.

\end{enumerate}

The cross-dataset configuration is deployed by taking the 50 known classes in the Aircraft dataset as the known classes, and taking all of the 200/196-class testing images in the CUB/Stanford-Cars dataset as the unknown-class images respectively.

For the coarse-grained setting, the proposed method is evaluated on the CIFAR+10/+50 and TinyImageNet datasets under a same data splitting manner as done in conventional OSR methods \cite{OpenHybrid, ARPL, MoEP-AE}. The split numbers of known/unknown classes are also listed in Table \ref{table: num_classes}.

\begin{enumerate}

	\item[-] \textbf{CIFAR+10/+50:} The CIFAR+10/+50 dataset is constructed from the two commonly-used datasets, \emph{i.e.}, CIFAR10 \cite{CIFAR10} and CIFAR100 \cite{CIFAR100}, which contain 10 and 100 categories of natural images, respectively. The whole 10 classes of the CIFAR10 dataset are used as the known classes, while 10 and 50 classes of the CIFAR100 dataset are chosen as the unknown classes for CIFAR+10 and CIFAR+50.
	
	\item[-] \textbf{TinyImageNet:} The TinyImageNet contains 200 categories of natural images, Here, 20 classes of the ImageNet dataset \cite{TinyImageNet} are chosen as the known classes, and non-overlapping 180 classes as the unknown classes.

\end{enumerate}

\subsubsection{Evaluation Metrics}

In this work, we adopt four evaluation metrics: the area under the receiver operating characteristic curve (AUROC) for evaluating the performance of open-set detection, the top-1 accuracy (ACC) for evaluating the performance of closed-set classification, the Open Set Classification Rate (OSCR) \cite{OSCR} for measuring the trade-off between accuracy and open-set detection rate, and the macro-F1 score for measuring the performance of both open-set detection and closed-set classification simultaneously. The first two metrics (\emph{i.e.}, AUROC and ACC) are used for the coarse-grained setting as done in \cite{OpenHybrid, OpenGAN, ARPL, MoEP-AE, OSR_transformer2}. The first three metrics {\emph{i.e.}, ACC, AUROC, and OSCR} are used for the fine-grained setting under the standard-dataset configuration as done in \cite{good-closed-set, OpenGAN, GMVAE-OSR, MoEP-AE}. And the fourth metric (\emph{i.e.}, macro-F1 score), which takes the unknown-classes as an additional class, is used for the fine-grained setting under the cross-dataset configuration as done in \cite{OpenGAN, IEEE-Access, MoEP-AE}.

%The ACC, AUROC, and OSCR metrics are used under the standard-dataset configuration in the fine-grained setting as done in , while the macro-F1 score is used under the cross-dataset configuration, as done in \cite{good-closed-set, OpenGAN, GMVAE-OSR, MoEP-AE}. For the coarse-grained setting, we use the same metrics (\emph{i.e.}, AUROC and ACC) as done in \cite{OpenHybrid, OpenGAN, ARPL, MoEP-AE, OSR_transformer2}.

\subsection{Implementation Details}

In our work, the image size in the CUB and Aircraft datasets is set to 448 $\times$ 448, and the image size in the rest datasets is set to 224 $\times$ 224. The Swin-B is used as the backbone, which is pretrained on the ImageNet-22K dataset. The AdamW \cite{AdamW} optimizer with weight decay of 0.01 and learning rate of $5\times10^{-5}$ is used for training the self-attention backbone module, and the SGD optimizer with weight decay of $1\times10^{-4}$ and initial learning rate of $1\times10^{-3}$ is used for training the rest parts of the network.

\begin{table}[t]
	\centering
	\caption{Comparison of the macro-F1 scores on the two fine-grained testing datasets (the Aircraft dataset is used for training, while the CUB and the Stanford-Cars datasets are used for testing respectively) under the cross-dataset configuration.}
	\begin{tabular}{m{3.3cm}<{\raggedright}m{1.5cm}<{\centering}m{2.1cm}<{\centering}}
		\toprule
		Method & CUB & Stanford-Cars \\
		\midrule
		OpenHybrid \cite{OpenHybrid} & 0.473 & 0.840 \\
		OpenGAN \cite{OpenGAN} & 0.426 & 0.819 \\	
		ARPL \cite{ARPL} & 0.469 & 0.831 \\
		GCPL \cite{GCPL} & 0.435 & 0.812  \\
		GMVAE-OSR \cite{GMVAE-OSR} & 0.454 & 0.826   \\
		CAMV \cite{IEEE-Access} & 0.461 & 0.833 \\	
		Cross-Entropy+ \cite{good-closed-set} & 0.482 & 0.869 \\
		\midrule
		Backbone & 0.497 & 0.886  \\
		OpenHybrid(Swin) & 0.499 & 0.875  \\
		ARPL(Swin) & 0.521 & 0.888  \\
		Cross-Entropy+(Swin) & 0.503 & 0.891  \\
		Trans-AUG \cite{OSR_transformer2} & 0.508 & 0.882 \\
		MoEP-AE-OSR \cite{MoEP-AE} & 0.529 & 0.876 \\
		STAN-OSFGR & \textbf{0.562} & \textbf{0.911} \\
		\bottomrule
	\end{tabular}
	\label{table: cross-dataset}
\end{table}

\subsection{Comparative Evaluation} \label{performance}

\subsubsection{Evaluation Under the Fine-Grained Setting}

As indicated in Sec. \ref{Related_Work} that only a few works (\emph{e.g.}, CAMV and Cross-Entropy+) have been evaluated on fine-grained datasets, in order to make a comprehensive comparison, we evaluate the proposed method in comparison to the referred CAMV and Cross-Entropy+, as well as seven state-of-the-art OSR methods (OpenHybrid, OpenGAN, ARPL, GCPL, GMVAE-OSR, MoEP-AE-OSR, and Trans-AUG) which have not been evaluated in their original papers. The corresponding results on the Aircraft, CUB, and Stanford-Cars datasets under the standard-dataset configuration are reported in Tables \ref{table: Aircraft}, \ref{table: CUB}, and \ref{table: Stanford-Cars}, respectively. The corresponding results under the cross-dataset configuration are reported in Table \ref{table: cross-dataset}. The results in these tables are split by a solid line according to transformer-based (swin transformer is used here) or non-transformer-based methods. The results of a vanilla transformer-based method used in Sec. \ref{Deficiency} (denoted as Backbone) are added for better comparison, and we also report the results of three state-of-the-art non-transformer-based methods by replacing their backbones with the swin transformer. Four points can be seen from these tables:

\begin{enumerate}
	\item[-] Firstly, STAN-OSFGR achieves the best performance under both the standard-dataset configuration and the cross-dataset configuration, and outperforms both the state-of-the-art OSR/OSFGR methods with the same backbone significantly in most cases, indicating that STAN-OSFGR can classify known-class fine-grained images accurately as well as detect unknown-class fine-grained images effectively.
	
	\item[-] Secondly, the results of transformer-based methods are universally better than those of non-transformer-based methods, indicating that the transformer is helpful for further improving the model performance.
	
	\item[-] Besides, the results of STAN-OSFGR are significantly better than those of Backbone, indicating that STAN improves the performance of the transformer backbone effectively when handling the OSFGR task.
	
	\item[-] Furthermore, the performance improvements on AUROC are more significant than those on ACC, and such improvements can be observed more obviously on the CUB and Aircraft datasets in a harder mode. These observations indicate that STAN has the ability for detecting semantic novelty, which is one of the most desired properties for handling the OSR task.
	
\end{enumerate}

\begin{figure}[t]
	\begin{center}
		\setlength{\abovecaptionskip}{0.cm}
		\includegraphics[height=6cm,width=8cm]{{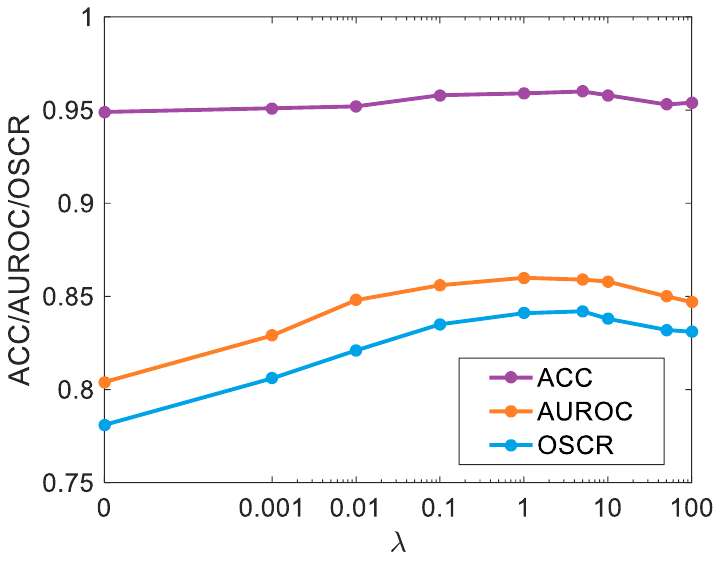}}
	\end{center}
	\caption{ACC, AUROC, and OSCR scores on the CUB dataset in the hard mode for different $\lambda$ in the total loss.}
	\label{fig: hyperparameter}
\end{figure} 

\begin{table*}[t]
	\centering
	\caption{Comparison of OSR results on the coarse-grained datasets (CIFAR+10/+50 and TinyImageNet).}
	\begin{tabular}{m{4.4cm}<{\raggedright}m{2.2cm}<{\centering}m{2.2cm}<{\centering}m{2.2cm}<{\centering}m{2.2cm}<{\centering}}
		\toprule
		Method & \multicolumn{2}{c}{CIFAR+10/+50} & \multicolumn{2}{c}{TinyImageNet}  \\
		\midrule
		Metric & ACC & AUROC & ACC & AUROC  \\
		\midrule
		OpenHybrid \cite{OpenHybrid} & 0.937 & 0.962/0.955 & 0.632 & 0.793 \\
		OpenGAN \cite{OpenGAN} & 0.940 & 0.981/0.983 & 0.684 & 0.907 \\	
		ARPL \cite{ARPL} & 0.940 & 0.965/0.943 & 0.638 & 0.762 \\
		GCPL \cite{GCPL} & 0.945 & 0.951/0.946 & 0.643 & 0.759   \\
		GMVAE-OSR \cite{GMVAE-OSR} & 0.952 & 0.952/0.947 & 0.729 & 0.782   \\
		CAMV \cite{IEEE-Access} & 0.942 & 0.942/0.933 & 0.796 & 0.805 \\	
		Cross-Entropy+ \cite{good-closed-set} & 0.958 & 0.954/0.939 & 0.862 & 0.826 \\
		\midrule
		Backbone & 0.981 & 0.959/0.963 & 0.904 & 0.909  \\
		%		STAN-Transformer & \textbf{0.989} & 0.982/0.984 & 0.951 & 0.918 \\
		%		Trans-ADH \cite{OSR_transformer1} & 0.996 & 0.994/0.990 & 0.959 & 0.977  \\
		OpenHybrid(Swin) & 0.981 & 0.978/0.980 & 0.925 & 0.934  \\
		ARPL(Swin) & 0.983 & 0.980/0.985 & 0.929 & 0.927   \\
		Cross-Entropy+(Swin) & 0.984 & 0.965/0.971 & 0.913 & 0.916   \\
		Trans-AUG \cite{OSR_transformer2} & 0.983 & 0.982/0.986 & 0.947 & 0.942  \\
		MoEP-AE-OSR \cite{MoEP-AE} & 0.983 & 0.976/0.975 & \textbf{0.965} & \textbf{0.952}  \\
		STAN-OSFGR & \textbf{0.988} & \textbf{0.994/0.991} & 0.953 & 0.945  \\
		\bottomrule
	\end{tabular}
	\label{table: coarse_grained_OSR}
\end{table*}

\begin{table*}[t]
	\centering
	\caption{Results on the CUB dataset by the proposed model with different configurations of the four modules.}
	\begin{tabular}{m{1.4cm}<{\centering}m{1.4cm}<{\centering}m{1.4cm}<{\centering}m{1.4cm}<{\centering}m{2.2cm}<{\centering}m{3.2cm}<{\centering}m{3.2cm}<{\centering}}
		\toprule
		Module1 & Module2 & Module3 & Module4 & ACC & \makecell[c]{AUROC\\ (Easy/Medium/Hard)} & \makecell[c]{OSCR\\ (Easy/Medium/Hard)} \\
		\midrule
		\ding{52} & \ding{56} & \ding{56} & \ding{56} & 0.949 & 0.945/0.875/0.804 & 0.908/0.848/0.781  \\
		\ding{52} & \ding{52} & \ding{56} & \ding{56} & 0.949 & 0.950/0.867/0.839 & 0.911/0.839/0.815  \\
		\ding{52} & \ding{52} & \ding{52} & \ding{56} & 0.956 & 0.962/0.917/0.848 & 0.928/0.891/0.829  \\
		\ding{52} & \ding{52} & \ding{52} & \ding{52} & \textbf{0.959} & \textbf{0.965/0.922/0.860} & \textbf{0.934/0.898/0.841}  \\
		\bottomrule
	\end{tabular}
	\label{table: ablation_study_modules}
\end{table*}

\subsubsection{Evaluation Under the Coarse-Grained Setting}

As the above experiments demonstrate, STAN-OSFGR is effective for discriminating fine-grained objects. Here, we further conduct experiments under the coarse-grained setting for investigating whether the STAN-OSFGR model could still be effective for discriminating coarse-grained objects. The corresponding results are reported in Table \ref{table: coarse_grained_OSR}. As seen from this table, STAN does not outperform other methods significantly and achieves similar performance to MoEP-AE-OSR, mainly because the transformer has learnt accurate attention maps for most of coarse-grained objects and these attentions are effective enough for discriminating coarse-grained images.

\subsection{Analysis of the Hyperparameter in Total Loss} \label{Hyperparameter}

Here, we analyze the influence of the hyperparameter $\lambda$ in Eqn. (3), which balances the weights of the spatial-feature classification loss $\mathcal{L}_{S}$ and the spatial-temporal-feature classification loss $\mathcal{L}_{ST}$ in the total loss function. The proposed STAN-OSFGR method is implemented with $\lambda=\{ 0, 0.001, 0.01, 0.1, 1, 5, 10, 50, 100 \}$ on the CUB dataset in the hard mode, and the testing results (ACC, AUROC, and OSCR) are reported in Fig. \ref{fig: hyperparameter}. As seen from this figure, the results vary slightly when $\lambda$ varies in $[ 0.1,10 ]$, demonstrating that the STAN-OSFGR method is not quite sensitive to the hyperparameter $\lambda$.

%\begin{table*}[htb]
%	\centering
%%	\setlength{\abovecaptionskip}{0pt}
%%	\setlength{\belowcaptionskip}{10pt}
%	\caption{Results on the CUB dataset by the proposed model with different numbers of layers in the two LSTMs.}
%	\begin{tabular}{m{2cm}<{\centering}m{2cm}<{\centering}m{1.5cm}<{\centering}m{4.5cm}<{\centering}m{4.5cm}<{\centering}}
%		\toprule
%		Outer LSTM & Inner LSTM & ACC & \makecell[c]{AUROC\\ (Easy/Medium/Hard)} & \makecell[c]{OSCR\\ (Easy/Medium/Hard)} \\
%		\midrule
%		2 & 1 & 0.951 & 0.959/0.901/0.825 & 0.921/0.871/0.802  \\
%		1 & 2 & 0.956 & 0.962/0.914/0.843 & 0.928/0.887/0.823  \\
%		2 & 2 & 0.953 & 0.960/0.898/0.820 & 0.923/0.869/0.799  \\
%		1 & 1 & \textbf{0.959} & \textbf{0.965/0.922/0.860} & \textbf{0.934/0.898/0.841}  \\
%		\bottomrule
%	\end{tabular}
%	\label{table: ablation_study_layers}
%	\vspace{-10pt}
%\end{table*}

\begin{table*}[t]
	\centering
	\caption{Results on the CUB dataset by the proposed model trained with different feature aggregation strategies.}
	\begin{tabular}{m{4.5cm}<{\raggedright}m{1.5cm}<{\centering}m{4.5cm}<{\centering}m{4.5cm}<{\centering}}
		\toprule
		Model & ACC & \makecell[c]{AUROC\\ (Easy/Medium/Hard)} & \makecell[c]{OSCR\\ (Easy/Medium/Hard)} \\
		\midrule
		Module1-AGG & 0.942 & 0.943/0.883/0.824 & 0.902/0.851/0.799  \\
		Module2-AGG & 0.949 & 0.950/0.867/0.839 & 0.911/0.839/0.815  \\
		Module3-AGG & 0.945 & 0.940/0.892/0.828 & 0.898/0.858/0.802  \\
		%		Reverse-AGG & 0.963 & 0.964/0.915/0.853 & 0.935/0.891/0.836  \\
		STAN-OSFGR & \textbf{0.959} & \textbf{0.965/0.922/0.860} & \textbf{0.934/0.898/0.841}  \\
		\bottomrule
	\end{tabular}
	\label{table: ablation_study_fusion}
\end{table*}

\begin{figure*}
	\begin{center}
		\setlength{\abovecaptionskip}{0.cm}
		\includegraphics[height=11cm,width=18cm]{{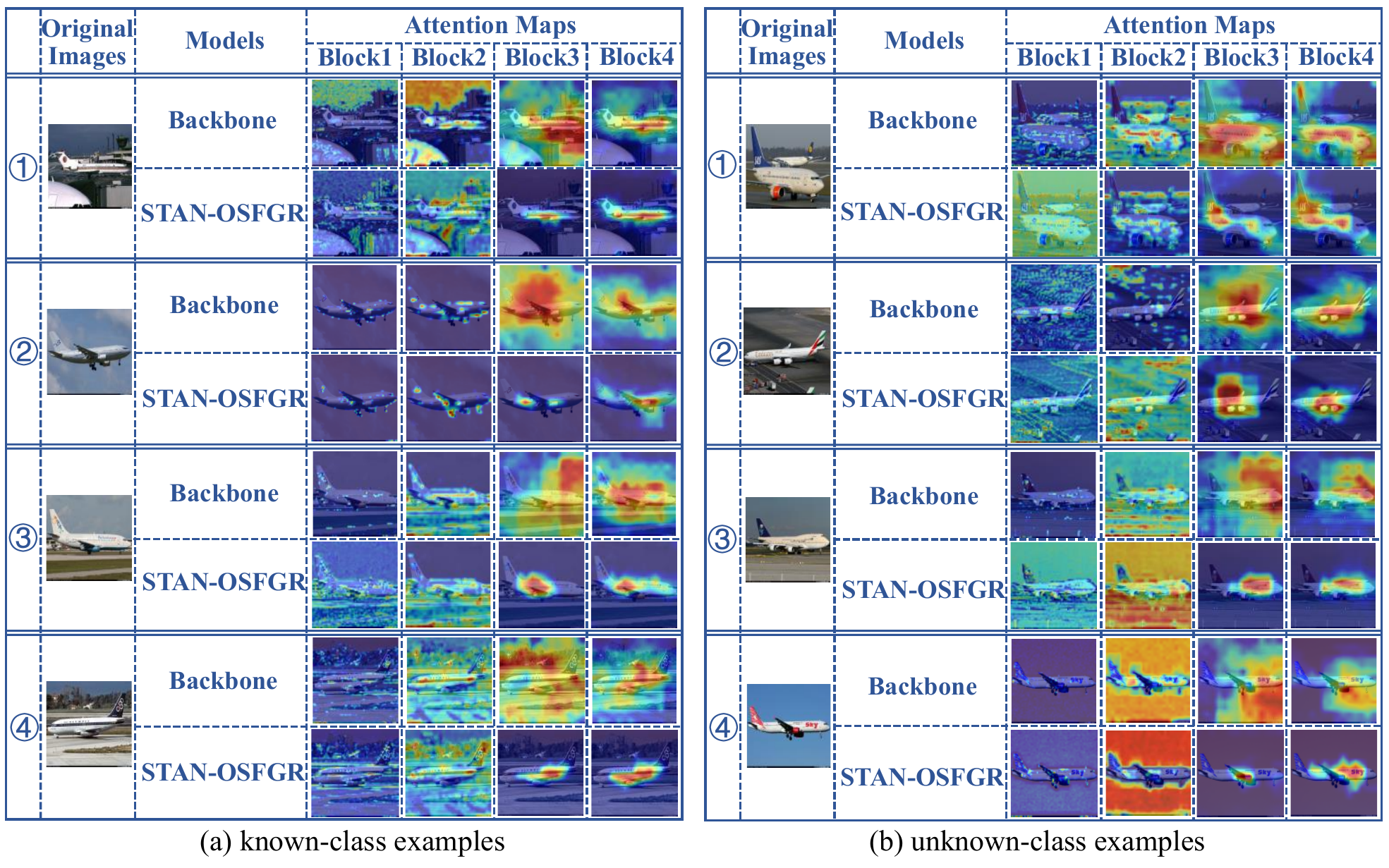}}
	\end{center}
%	\vspace{-10pt}
	\caption{Visualization of the attention maps learnt by both the backbone method (\emph{i.e.}, the swin transformer) and the proposed method for (a) the four known-class images and (b) the four unknown-class images from the Aircraft dataset. All of these images are correctly predicted by STAN-OSFGR. The attentions in red regions are strongest, while the attentions in yellow, green, and blue regions decrease by degrees.}
	%	It can be seen that the attentions are more accurate on the object with Module3 added, and they are relatively more complete with Module 4 added in the model.}
	\label{fig: train_attentions}
%	\vspace{-10pt}
\end{figure*} 

\subsection{Ablation Studies} \label{ablation studies}

Here, we conduct two ablation studies on modules and feature aggregation strategies, respectively. The experiments are all conducted on the CUB dataset.

\subsubsection{Ablation Study on the Modules}

Firstly, we conduct an ablation study on modules for evaluating the effect of the three explored modules in STAN (\emph{i.e.}, the SFSO module, the STFL module, and the CA module) which are introduced in Sec. \ref{STAN-modules}, and the results are reported in Table \ref{table: ablation_study_modules}. The model only with Module1 represents the Backbone model, the model only with Module1 and Module2 aggregates the sequential features obtained from the SFSO module and uses the aggregated features for discriminating fine-grained objects, the model only with Module1, Module2 and Module3 removes the CA module in STAN-OSFGR, and the model with all of the four modules represents the STAN-OSFGR model. As Table \ref{table: ablation_study_modules} shows, all of the three explored modules (the SFSO module, the STFL module, and the CA module) play important roles for handling the OSFGR task, among which the STFL module (\emph{i.e.}, the spatial-temporal feature learning module) contributes significant improvement and the CA module (\emph{i.e.}, the context-aware module) further improves the performance especially on detecting unknown classes in the `Hard' mode. These observations indicate that modeling the temporal dependency improves the performance of Backbone for handling the OSFGR task uniformly and significantly, and using the contextual information further improves the ability of the model for discriminating fine-grained objects.

\subsubsection{Ablation Study on the Feature Aggregation Strategies}

Considering that a temporal aggregation strategy is used in STAN-OSFGR, hence we also conduct another ablation study for comparing against three models that are configured with three different feature aggregation strategies based on STAN: Module1-AGG where the spatial features outputted from the self-attention backbone module are straightforwardly concatenated, Module2-AGG where the sequential features outputted from the SFSO module are concatenated, and Module3-AGG where the features outputted at all moments from the STFL module are concatenated.
%Additionally, we also compare against another feature aggregation strategy which is similar to the aggregation strategy in STAN-OSFGR but reverses the feature sequence for analyzing the effect of the order, and the trained model is denoted as Reverse-AGG. 
The corresponding results are reported in Table \ref{table: ablation_study_fusion}. Two points can be seen from this table:

\begin{itemize}
	
	\item[-] STAN-OSFGR achieves significantly better performance than Module1-AGG and Module2-AGG, indicating the importance of temporal aggregation.
	
	\item[-] The results of STAN-OSFGR are better than those of Module3-AGG, because the aggregation of earlier outputs weakens the effect of the temporal aggregation. 
	
\end{itemize}

\subsection{Visualization} \label{visualization}

In order to analyze whether the attention maps learnt by the network have been improved, we take the eight examples from the Aircraft dataset (which have been used in Sec. \ref{Deficiency} and all of these images are predicted correctly by STAN-OSFGR) as examples, and the corresponding attention maps learnt by two models (Backbone and STAN-OSFGR) are visualized in Fig. \ref{fig: train_attentions}. As seen from this figure, the attention maps learnt by the higher blocks of the STAN-OSFGR model focus on fine-grained objects more accurately than those learnt by the Backbone model.

%\begin{itemize}
%
%	\item[-] The final attentions learnt by both STAN-OSFGR w/o the CA module and STAN-OSFGR focus on fine-grained objects more accurately than those learnt by Backbone, due to the modeled temporal dependency.
%	
%	\item[-] The final attentions learnt by STAN-OSFGR cover the regions of fine-grained objects more completely than those learnt by STAN-OSFGR w/o the CA module, because the CA module (\emph{i.e.}, the context-aware module) makes use of contextual information when the temporal dependency is modeled by the STFL module.
%	
%\end{itemize}

\section{Conclusion}   \label{Conclusion}

In this paper, we firstly find that the vision transformer could sometimes not learn accurate attention maps for discriminating fine-grained objects. Addressing this observation, we propose the spatial-temporal attention network, STAN, which performs four modules for learning accurate attentions and fine-grained feature representations: the self-attention module, the spatial feature self-organizing module, the spatial-temporal feature learning module, and the context-aware module. Moreover, the STAN-OSFGR method is designed for handling the OSFGR task with our proposed STAN. Experimental results on both the 3 fine-grained OSR datasets and 2 coarse-grained OSR datasets demonstrate the effectiveness of STAN-OSFGR for handling the OSR tasks, especially for handling the OSFGR task. 

In the future, we will apply the proposed STAN network in other computer vision tasks where the attention mechanisms play essential roles, such as semantic segmentation, object detection, and monocular depth estimation for learning more effective attention maps as well as feature representations.

\ifCLASSOPTIONcaptionsoff
  \newpage
\fi

%\newpage

\small
\bibliographystyle{IEEEtran}
%\bibliography{egbib_20220614}
\bibliography{egbib_20220622}

%~\\
%
%~\\
%
%\noindent \textbf{Jiayin Sun} is pursuing the Ph.D. in pattern recognition and intelligence systems with the National Laboratory of Pattern Recognition, Institute of Automation, Chinese Academy of Sciences. Her current research interests include machine learning and its applications, particularly in open-set recognition.\par
%~\\
%\noindent \textbf{Qiulei Dong} received the B.S. degree in automation from Northeastern University in 2003 and the Ph.D. degree from the Institute of Automation, Chinese Academy of Sciences in 2008. He is currently a Professor with the National Laboratory of Pattern Recognition, Institute of Automation, Chinese Academy of Sciences. His current research interests include 3-D computer vision and pattern classification.\par

\end{document}